\pdfoutput=1

\documentclass[11pt]{article}

\usepackage[preprint]{acl}

\usepackage{times}
\usepackage{latexsym}
\usepackage{amsmath}

\usepackage[T1]{fontenc}

\usepackage[utf8]{inputenc}

\usepackage{microtype}

\usepackage{inconsolata}

\usepackage{graphicx}

\usepackage{microtype}
\usepackage{subfigure}
\usepackage{booktabs} 
\usepackage{wrapfig}
\usepackage{cuted}
\usepackage{lipsum}
\usepackage{midfloat}
\usepackage{lmodern,babel,adjustbox,booktabs,multirow}
\usepackage{threeparttable}
\usepackage{subcaption}
\usepackage{comment}
\usepackage{array, makecell}

\title{Concepts or Skills? Rethinking Instruction Selection for Multi-modal Models}

\author{\textbf{Andrew Bai}, \textbf{Justin Cui}, \textbf{Ruochen Wang}, \textbf{Cho-Jui Hsieh} \\ \\
    Department of Computer Science \\
  University of California, Los Angeles
  }

\begin{document}
\maketitle
\begin{abstract}
Vision-language instruction tuning achieves two main purposes: learning visual concepts and learning visual skills.
In this paper, we found that vision-language benchmarks fall into the dichotomy of mainly benefiting from training on instructions with similar skills or visual concepts.
Inspired by the discovery, we designed a simple targeted training data selection method to optimize the performance of a given benchmark.
We first extract the concepts/skills from the benchmark, determine whether the benchmark predominantly benefits from similar concepts or skills, and finally select instructions with the most matching concepts/skills.
Experiments on 10+ benchmarks validate the effectiveness of our targeted data selection method, showing +0.9\% over the best existing baseline averaged over all benchmarks and +1.5\% on the skill-focused subset.
Our findings underscore the importance of recognizing the inherent trade-off within instruction selection, which requires balancing the acquisition of conceptual knowledge against visual skill.
\end{abstract}

\section{Introduction}


Recent progress in vision-language modeling has been driven by the modular combination of large pretrained language models (LLMs) with powerful visual encoders, typically connected via a modality adapter that transforms visual features into a format compatible with the LLM’s input space. 
This architecture allows models to leverage the linguistic capabilities of LLMs while incorporating rich visual understanding from pretrained vision backbones. 
However, simply joining these components is insufficient for strong multimodal performance; extensive continual pretraining on paired vision-language data is often required to bridge the modality gap. 
In addition, vision-language instruction tuning—where the model is further trained on curated prompts that teach specific multimodal capabilities—has become essential for aligning the joint model’s behavior with desired visual-linguistic tasks.

Vision-language instruction tuning has emerged as a powerful framework for aligning multimodal models with human-desired behaviors, especially in tasks requiring both visual perception and linguistic reasoning. 
This tuning process serves two major purposes: first, it helps models learn to associate visual representations with the corresponding textual concepts, and second, it enables the acquisition of new visual capabilities such as counting objects, reasoning about spatial relations, and inferring physical properties. 
These dual roles are critical for deploying vision-language models in real-world applications where generalization to both seen and unseen instructions is required.

Yet, despite the widespread adoption of instruction tuning, the vision-language benchmarks used for evaluation often vary in the types of capabilities they measure. 
Some emphasize concept recognition (e.g., visual question answering), while others emphasize reasoning or skill-based tasks (e.g., counting, relational understanding, or commonsense inference). 
This discrepancy raises a fundamental question: Should instruction selection prioritize alignment with the skills required by a task or the concepts present in the visual content? 
More concretely, can performance on a benchmark be improved by tuning on a dataset that emphasizes similar skills versus one that emphasizes similar visual concepts?

To investigate this, we conduct a systematic analysis across a diverse range of vision-language benchmarks, aiming to understand how different types of instruction similarity influence downstream performance. 
Our study reveals that existing benchmarks fall into two distinct categories: those that benefit more from instructions involving similar skills (e.g., object counting across different domains) and those that benefit more from instructions involving similar concepts (e.g., object or scene categories common to the benchmark). 
Notably, this pattern persists across architectures and training regimes, suggesting that the division is intrinsic to the nature of the tasks and not merely an artifact of the model design.

Motivated by these findings, we propose a simple yet effective targeted instruction selection method that adapts to the nature of each benchmark. 
Specifically, we first extract the dominant concepts and skills present in a given evaluation set using automated instruction parsing and skill/concept taxonomy alignment. 
We then determine whether the benchmark is concept-dominant or skill-dominant using validation performance differentials, and finally, we select training instructions that most closely align with the benchmark’s dominant type. 
This targeted strategy allows the model to focus on the most relevant inductive biases, whether conceptual or procedural.

We evaluate our approach across more than ten standard vision-language benchmarks, covering a wide range of task types and difficulty levels. 
Our results demonstrate consistent performance gains, with an average improvement of 0.9\% over the strongest baseline using untargeted instruction tuning. 
These findings underscore the importance of benchmark-aware instruction selection and open new directions for task-adaptive multimodal learning.

\section{Related Work}
\subsection{Vision Language Model}
Large language models (LLMs)\cite{dubey2024llama,openaiOpenAIPlatform,team2024gemini, chiang2023vicuna} have demonstrated remarkable performance across a wide variety of tasks. This success is largely attributed to pretraining on trillions of tokens, followed by post-training techniques such as reinforcement learning with human feedback (RLHF)\cite{christiano2017deep}. Building on their capabilities in the text domain, recent research has extended LLMs to handle additional modalities such as images. MiniGPT-4\cite{zhu2023minigpt} integrates a pretrained Vision Transformer (ViT) backbone with a Q-Former and a single linear projection layer, combining it with the Vicuna language model to achieve strong performance on multimodal tasks. Concurrently, InstructBLIP\cite{dai2023instructblip} employs a similar approach to generate instruction-following responses conditioned on both images and text prompts. LLaVA~\cite{liu2023visual, liu2024improved} converts a large language model into a multimodal model by first encoding images with a CLIP~\cite{radford2021learning} encoder, then mapping the visual features into the text embedding space through a linear MLP, forming a simple yet effective integration strategy.

\subsection{Visual Instruction Data Selection}
Dataset selection~\cite{har2004coresets, roux2012stochastic, campbell2018bayesian, mirzasoleiman2020coresets} has been extensively explored to enhance the training efficiency of models. Recently, this line of research has been extended to vision-language models, aiming to reduce training costs while preserving performance. Coincide~\cite{lee2024concept} partitions the dataset into numerous subsets and retains samples based on the transferability of clusters. 
ICONS~\cite{wu2024icons} identifies important samples by measuring the influence of individual data points, defined via gradient similarity with a validation set.
Prism~\cite{bi2025prism} introduces a training-free method that utilizes Pearson correlation analysis to measure the intrinsic visual encoding capabilities of MLLMs. In contrast, PreSel~\cite{safaei2025filter} approaches the problem differently: it first applies a filtering mechanism to identify high-quality images, and only then generates instructions for those selected samples. Despite significant progress in achieving strong performance with reduced data, there is limited research on how different tasks are affected by underlying concepts or skills.

\subsection{Concepts vs Skills}
We distinguish sharply between the dichotomy: \emph{concepts}, which refer to the visual entities, attributes, and objects present in an image, i.e., \textbf{what} appears, and \emph{skills}, which encapsulate the reasoning operations or judgment strategies necessary for correctly interpreting or answering questions about those entities, i.e., \textbf{how} to analyze them). 
This distinction echoes the formal view of compositionality in VQA proposed by \citet{whitehead2021separatingskillsconceptsnovel}, who explicitly model a skill–concept decomposition to enable generalization across unseen combinations of skills (e.g. ``color'' judgment) and concepts (e.g. ``car''). 
While concepts ground models in visual content, skills capture the latent reasoning patterns—such as counting, spatial inference, or trend analysis—that drive downstream task success. 
Coincide~\citep{lee2024concept} integrates concept and skill to jointly define a notion of similarity, which is then used to maximize diversity in the selected samples.
In our case, decoupling these two axes allows selection methods to target training examples based on what a model needs to recognize versus what it needs to reason about, facilitating more precise alignment with the cognitive demands of various vision–language benchmarks.

\section{Methodology}

\subsection{Problem Formulation}

Let $\mathcal{I} = \{(x_i, y_i)\}$ denote a pool of multi-modal instruction examples, where each $x_i$ is a natural language instruction paired with one or more images, and $y_i$ is the expected response. Given a target set of downstream tasks $\mathcal{T} = \{T_1, ..., T_K\}$, our objective is to select a subset $\mathcal{I}^* \subset \mathcal{I}$ that maximizes performance on the downstream tasks after instruction tuning. Each task $T_k$ corresponds to a benchmark dataset characterized by a specific input-output format and evaluation metric. We hypothesize that the most beneficial instruction subsets vary by task, and that alignment between the instructions’ visual content and the task’s demands plays a crucial role in downstream generalization.

\section{Methodology}

This study introduces a retrieval-based framework to compare \emph{concept-prioritized} and \emph{skill-prioritized} selection strategies for curating vision-language instruction-tuning data. The approach explicitly separates the notion of \emph{visual concept} from that of \emph{visual skill}. For a given instruction, similar examples are retrieved from a candidate pool using nearest-neighbor search performed either in the concept space or in the skill space.

\subsection{Concept Representation}
The \emph{concept representation} is derived from the image associated with each instruction. Images are passed through a pretrained vision encoder, and the resulting embeddings are used to characterize the visual content. This representation captures semantic similarity between images while remaining agnostic to linguistic information.

\subsection{Skill Representation}
The main technical contribution of this work lies in the construction of a \emph{skill representation}. 
Unlike concepts, visual skills are not directly annotated in datasets and must be inferred. To address this, an automated pipeline is developed to extract and encode skill information for each instruction:

\begin{enumerate}
    \item \textbf{Skill isolation through large language models} For each instruction--image--answer triplet, a large language model is prompted with the question: \textit{``What visual skills are required to answer this instruction correctly?''}. The response consists of a concise list of skills, such as object counting, spatial reasoning, or fine-grained attribute recognition.
    \item \textbf{Skill embedding extraction} The skill descriptions generated by the language model are converted into fixed-dimensional vectors using a pretrained sentence embedding model. These embeddings define a skill space in which instructions with similar cognitive requirements are close to each other.
\end{enumerate}

This process yields a representation that explicitly decouples \emph{what an image depicts} from \emph{the reasoning skills required} to interpret it.

\subsection{Nearest-Neighbor Data Selection}
Given a query instruction, nearest-neighbor retrieval is applied in either the concept space or the skill space.
\textbf{Concept-prioritized selection} retrieves examples with the closest concept embeddings.
\textbf{Skill-prioritized selection} retrieves examples with the closest skill embeddings.
Two distinct subsets of data are thereby produced, differing only in the selection criterion.

\subsection{Downstream Evaluation}
The curated datasets are subsequently used for instruction tuning, enabling a direct comparison of the effect of concept-driven versus skill-driven data selection on downstream multimodal performance.
Specifically for a given benchmark, the model trained on its concept neighbors is denoted as ``Concept$\uparrow$'' and skill neighbors as ``Skill$\uparrow$''.

\section{Experiments}

We evaluate our proposed instruction selection strategies across twelve diverse vision-language benchmarks and two instruction datasets. Our goal is to measure the effectiveness of concept- and skill-targeted instruction selection in comparison with several untargeted baselines under data budget constraints.

\subsection{Experimental Setup}

\paragraph{Dataset} We conduct training budget constraint experiments on two datasets: LLaVA-1.5~\citep{liu2023improvedllava} with 665k examples and ALLaVA-4V~\citep{chen2024allava} with 1.2M examples. 
Each example includes one or more instruction-response pairs and an associated image.
We experimented with sampling budgets of 5\% and 10\% for the LLaVA-1.5 instruction pool and 2.5\% for the ALLaVA-4V instruction pool.

\paragraph{Baselines} We compare concept-prioritized (\texttt{Concept$\uparrow$}) and skill-prioritized (\texttt{Skill$\uparrow$}) targeted selection strategies against three untargeted baselines: \textbf{Random} sampling, \textbf{Coincide}~\citep{lee2024concept} which selects samples maximizing diversity, and \textbf{PreSel}~\citep{safaei2025filter} which only relies on the image for instruction selection.
Notably, although ICONS~\citep{wu2024icons} is also a relevant baseline for targeted selection, we were unable to reproduce their results due to the high computation cost associated with caching the LoRA gradients for every training instance.

\paragraph{Training details} We implement instruction tuning using the LLaVA-1.5 framework, which integrates a CLIP-ViT image encoder with a Vicuna-7B language model. Fine-tuning is performed using LoRA adapters for parameter efficiency. All models are trained for one epoch using the AdamW optimizer with a learning rate of $2 \times 10^{-5}$, a batch size of 128, and an image resolution of $224 \times 224$ pixels. Training is distributed across four NVIDIA A6000 GPUs.
Performance for methods with reported standard deviation is averaged over three random seeds.

\paragraph{Embedding details} We precompute embeddings for all instruction examples and benchmark samples. 
We use the FAISS library for efficient nearest neighbor search to identify top-k most similar instruction samples per benchmark. 
Concept embeddings are obtained using a zero-shot CLIP image feature extractor on the images.
Skill embeddings are obtained using first prompting GPT-4o with the questions in an instruction for the relevant visual skills required to answer the question. 
The skill descriptions are then encoded with an open-sourced sentence transformer, \href{https://huggingface.co/sentence-transformers/all-MiniLM-L6-v2}{MiniLM-L6-v2}, to extract a 384-dimensional dense embedding.

\subsection{Evaluation Protocol}

We evaluate each model in a zero-shot setting on the downstream benchmarks, reporting task-specific metrics such as accuracy (GQA, ScienceQA) and exact match (TextVQA, OCR-VQA) where applicable. 
The random baseline is repeated with three random seeds to account for training variability, and we report the average performance along with standard deviations.
For other baselines, we report the results of one experiment run due to computation constraints.
The standard deviations from the random baseline are utilized to determine if other baselines outperform random statistically significantly.

We evaluate our methods on twelve benchmarks: VQAv2, GQA, VizWiz, ScienceQA (SQA-I), TextVQA, POPE, MME, MMBench (en), LLaVA-Bench, AI2D, OK-VQA, and ST-VQA. These benchmarks encompass a variety of tasks, including general VQA, OCR, and scientific reasoning. 

\subsection{Experiment Results}
Table~\ref{tab:llava_5pp},~\ref{tab:llava_10pp}, and~\ref{tab:allava_2.5pp} present the comparisons of the baselines on different benchmarks.
The benchmarks are split into three sections separated by horizontal lines: the top section corresponds to concept-targeted tasks, while the bottom section corresponds to skill-targeted ones.
The middle section consists of benchmarks where the best performing method does not outperform the average of the random uniform baseline plus its standard deviation.
The best performing method is highlighted in bold, and the second is underlined.

\paragraph{LLaVA 5\%}
Table~\ref{tab:llava_5pp} presents results under the 5\% LLaVA budget. In this low-resource setting, instruction selection is particularly impactful. Targeted strategies yield strong gains on most benchmarks, especially the skill-focused tasks with specific domain characteristics or reasoning requirements.

\begin{table*}[!ht]
    \centering
    \caption{Experiment results for 5\% data selection on LLaVA-1.5.
    }
    \adjustbox{max width=0.9\linewidth}{
    \begin{threeparttable}
    \begin{tabular}{lrrrrrrrrr}
        \toprule
            Benchmark &  \multicolumn{3}{c}{Untargeted} & \multicolumn{3}{c}{Targeted (Ours)} \\
             & Random & Coincide & PreSel & Concept$\uparrow$ & Skill$\uparrow$ & C-S\\
        \midrule
        \midrule
        VizWiz      & 28.4 $\pm$ 0.7 & \underline{28.7} & 28.4 & \textbf{30.2} & 28.6 & +1.6\\   
        LlaVa-Bench & 66.7 $\pm$ 1.4 & \underline{68.0} & \textbf{68.4} & \underline{68.0} & 67.4 & +0.6 \\   
        VQAV2       & 72.2 $\pm$ 0.2 & \textbf{73.3} & \underline{72.5} & 72.1 & 71.6 & +0.5\\   
        TextVQA     & 52.0 $\pm$ 0.3 & 52.1 & \underline{51.1} & \textbf{54.8} & 54.4 & +0.4\\   
        GQA         & 52.7 $\pm$ 0.5 & 53.6 & 52.0 & \textbf{54.1} & \underline{54.0} & +0.1\\   
        MME         &1259.3 $\pm$ 12.8 &\textbf{1343.0}&\underline{1326.3}&1302.4&1248.2&+54.2\\   
        \midrule
        MMBench(en) & 55.8 $\pm$ 1.4 & 55.2 & \underline{57.0} & \textbf{57.1} & \underline{57.0} & +0.1\\   
        POPE        & 84.3 $\pm$ 0.6 & 83.9 & \underline{84.4} & 83.5 & \textbf{84.7} & -1.2\\   
        \midrule
        STVQA       & 44.7 $\pm$ 0.1 & 46.5 & 45.1 & \underline{46.8} & \textbf{47.5} & -0.7\\   
        SQA-I       & 65.9 $\pm$ 0.5 & 66.3 & \underline{66.7} & 65.8 & \textbf{68.7} & -2.9\\   
        AI2D        & 50.8 $\pm$ 0.6 & 50.5 & \underline{51.1} & 49.1 & \textbf{53.8} & -4.7\\   
        OK-VQA      & 45.2 $\pm$ 0.7 & \textbf{51.7} & 42.4 & 43.2 & \underline{50.8} & -7.6\\   
        \bottomrule
        \end{tabular}
        \end{threeparttable}
        }
    \label{tab:llava_5pp}
\end{table*}

\paragraph{LLaVA 10\%}
Table~\ref{tab:llava_10pp} shows results with a 10\% budget. While increased data volume reduces the performance gap between methods, targeted selection still provides notable improvements on benchmarks with strong domain or skill alignment.
Notably more benchmarks benefit from skill-focused selection, indicating that the signal saturates more easily with concept-focused selection.

\begin{table*}[!ht]
    \centering
    \caption{Experiment results for 10\% data selection on LLaVA-1.5.
    }
    \adjustbox{max width=0.9\linewidth}{
    \begin{threeparttable}
    \begin{tabular}{lrrrrrrrr}
        \toprule
            Benchmark &  \multicolumn{3}{c}{Untargeted} & \multicolumn{3}{c}{Targeted (Ours)} \\
             & Random & Coincide & PreSel & Concept$\uparrow$ & Skill$\uparrow$ & C-S \\
        \midrule
        \midrule
        VizWiz      & 28.8 $\pm$ 1.1 & 29.7 & \underline{29.8} & 29.2 & \textbf{30.8} & -1.6\\   
        LlaVa-Bench & 67.0 $\pm$ 3.2 & \textbf{68.6} & 66.9 & \underline{68.4} & 68.2 & +0.2\\   
        VQAV2       & 74.0 $\pm$ 0.2 & \textbf{75.0} & 74.0 & \underline{74.5} & 73.6 & +0.9\\   
        TextVQA     & 53.5 $\pm$ 0.7 & 53.4 & 53.3 & \underline{55.0} & \textbf{55.7} & -0.7\\   
        GQA         & 56.0 $\pm$ 0.2 & \underline{56.6} & 56.0 & \underline{56.6} & \textbf{57.2} & -0.6 \\   
        MME         &1349.6$\pm$ 34.1 &\underline{1382.4}&\textbf{1387.3}&1368.6&1302.7 & +65.9\\   
        \midrule
        MMBench(en) & 58.1 $\pm$ 0.8 & \textbf{60.8} & 57.7 & 57.7 & \underline{59.5} & -1.8\\   
        POPE        & 84.0 $\pm$ 1.0 & \underline{84.3} & \underline{84.3} & \textbf{84.9} & \underline{84.3} & -0.6 \\   
        \midrule
        STVQA       & 47.1 $\pm$ 0.6 & \underline{48.2} & 47.9 & 47.7 & \textbf{48.7} & -1.0 \\   
        SQA-I       & \underline{67.7 $\pm$ 0.5} & 66.9 & 66.0 & 67.1 & \textbf{70.4} & -3.3\\   
        AI2D        & 52.2 $\pm$ 0.5 & 53.3 & 51.5 & 51.9 & \textbf{54.2} & -2.3\\   
        OK-VQA      & 49.5 $\pm$ 1.3 & \textbf{53.7} & 50.4 & 50.7 & \underline{51.9} & -1.2 \\   
        \bottomrule
        \end{tabular}
        \end{threeparttable}
        }
    \label{tab:llava_10pp}
\end{table*}

\paragraph{ALLaVA 2.5\%}
Table~\ref{tab:allava_2.5pp} reports results from the 2.5\% ALLaVA setting, where instructions are more diverse and noisier. 
In this setting, the targeted baselines significantly outperform the random baseline, highlighting the importance of instruction quality and alignment in low-data regimes, especially for larger and more diverse instruction datasets.

\begin{table}[h!]
    \centering
    \caption{Results for ALLaVA 2.5\% data selection.
    }
    \adjustbox{max width=\linewidth}{
    \begin{threeparttable}
    \begin{tabular}{lrrrrrr}
        \toprule
            Benchmark &  \multicolumn{1}{c}{Untargeted} & \multicolumn{3}{c}{Targeted (Ours)} \\
             & Random & Concept$\uparrow$ & Skill$\uparrow$ & C-S \\
        \midrule
        \midrule
        VizWiz      & 21.0 $\pm$ 0.4 & \textbf{31.1} & 30.5 & +0.6\\   
        LlaVa-Bench & 63.2 $\pm$ 2.5 & \textbf{76.6} & 70.7 & +5.9\\   
        VQAV2       & 52.5 $\pm$ 2.8 & \textbf{72.3} & 70.9 & +1.4\\   
        TextVQA     & 36.6 $\pm$ 3.2 & 52.2 & \textbf{52.8} & -0.6\\   
        GQA         & 34.8 $\pm$ 1.9 & \textbf{52.7} & 51.7 & +1.0\\   
        MME         & 854.4 $\pm$ 97.8 &1208.5&\textbf{1222.9}& -14.4\\   
        \midrule
        MMBench(en) & 29.9 $\pm$ 1.8 & 52.0 & \textbf{54.8} & -2.8\\   
        POPE        & 76.1 $\pm$ 1.6 & \textbf{83.3} & 82.3 & +1.0\\   
        \midrule
        STVQA       & 29.8 $\pm$ 3.5 & 46.8 & \textbf{47.5} & -0.7\\   
        SQA-I       & 44.9 $\pm$ 5.4 & 62.1 & \textbf{66.5} & -4.4\\   
        AI2D        & 42.4 $\pm$ 2.6 & 50.3 & \textbf{54.0} & -3.7\\   
        OK-VQA      &  0.4 $\pm$ 0.3 & \textbf{38.4} & 24.7 & +13.7\\   
        \bottomrule
        \end{tabular}
        \end{threeparttable}
        }
    \label{tab:allava_2.5pp}
\end{table}

\section{Discussion}

\subsection{Different Prioritization Benefits Different Downstream Tasks}

Our results indicate that the effectiveness of data selection strategies depends strongly on the nature of the benchmark. \emph{Concept-prioritized selection} tends to benefit benchmarks where success depends primarily on recognizing and localizing objects within an image. These tasks, such as VQAv2, VizWiz, and MME, are dominated by relatively straightforward yes/no or short-answer questions that can be answered once the relevant visual elements are correctly identified. In contrast, \emph{skill-prioritized selection} shows clear advantages on benchmarks that demand reasoning or specialized visual competencies beyond object recognition. Datasets such as SQA-I, OK-VQA, and AI2D require abilities like commonsense reasoning, fine-grained attribute discrimination, reading embedded text (OCR), or multi-step inference over visual evidence. The divergence in performance suggests that concept-driven and skill-driven selection are complementary: the former strengthens a model's ability to ground answers in visual content, while the latter enhances its ability to execute more complex reasoning over that content.


\subsection{Untargeted Baselines Implicitly Trade Off Performance}

While untargeted baselines such as Coincide and PreSel perform well on average, they tend to make implicit tradeoffs across tasks. 
These methods favor globally frequent instruction patterns, which may benefit general-purpose benchmarks like VQAv2 or GQA but degrade performance on domain-specific or skill-intensive tasks. 
Several concept-centric benchmarks are not reported in the original baseline studies, and we observe that performance on such unreported tasks tends to suffer when these heuristics are applied. 
This highlights the importance of understanding the hidden biases and tradeoffs that come with untargeted selection. Another consequence of this global-averaging approach is that special cases such as TextVQA or SQA, which require more specialized skills (e.g., OCR, compositional reasoning), are easily underrepresented and neglected. Our findings suggest that incorporating a stronger focus on skill diversity into otherwise untargeted selection strategies may help mitigate these limitations, reducing the cost of optimizing for average performance while still supporting tasks that fall outside the dominant distribution.

\subsection{Hybrid Selection Strategies}

We also investigated whether combining concept-targeted and skill-targeted strategies could provide the best of both worlds. 
For each benchmark, we computed a relevance score for each instruction with respect to visual concepts and skills, and then explored multiple methods for combining these scores into a unified selection criterion. 
Specifically, we experimented with (1) summing the concept and skill relevance scores, (2) taking the maximum of the two scores, and (3) splitting the selection budget evenly so that half of the instructions were chosen based on concept relevance and the other half on skill relevance.

Table~\ref{tab:llava_5pp_hybrid} presents the three hybrid approaches on a representative concept- and skill-targeted subtasks. 
Surprisingly, none of these hybrid strategies consistently outperformed the single-targeted approaches. 
In nearly all benchmarks, the worst hybrid variant underperformed the better of the two targeted strategies.
This suggests that combining the two signals indiscriminately dilutes the effect of the dominant alignment factor, whether it be concept or skill, and there is no straightforward method to directly incorporate the concept and skill signals.
Our findings imply that benchmarks tend to benefit strongly from one type of alignment at a time rather than a mixture of both.

\begin{table}[h!]
    \centering
    \caption{Experiment results for 5\% data selection on LLaVA-1.5 for comparing hybrid strategies for incorporating concept and skill signals.
    }
    \adjustbox{max width=\linewidth}{
    \begin{threeparttable}
    \begin{tabular}{lrrrrr}
        \toprule
            Benchmark & \multicolumn{5}{c}{Concept-Skill Hybrids} \\
             & Max & Sum & Split & Concept$\uparrow$ & Skill$\uparrow$ \\
        \midrule
        \midrule
        GQA         & 53.9 & \textbf{54.9} & 53.7 & \underline{54.1} & 54.0\\   
        MME         &1312.3&\underline{1312.5}&\textbf{1337.2}&1302.4&1248.2\\   
        \midrule
        SQA-I       & \textbf{70.1} & \underline{69.3} & 68.1 & 65.8 & 68.7\\   
        OK-VQA      & 41.9 & \underline{47.4} & \underline{47.4} & 46.8 & \textbf{47.5}\\
        \bottomrule
        \end{tabular}
        \end{threeparttable}
        }
    \label{tab:llava_5pp_hybrid}
\end{table}

\subsection{Predicting Benchmark Alignment via Mutual Ranking}

\begin{figure}[ht]
  \centering
  \includegraphics[width=0.9\linewidth]{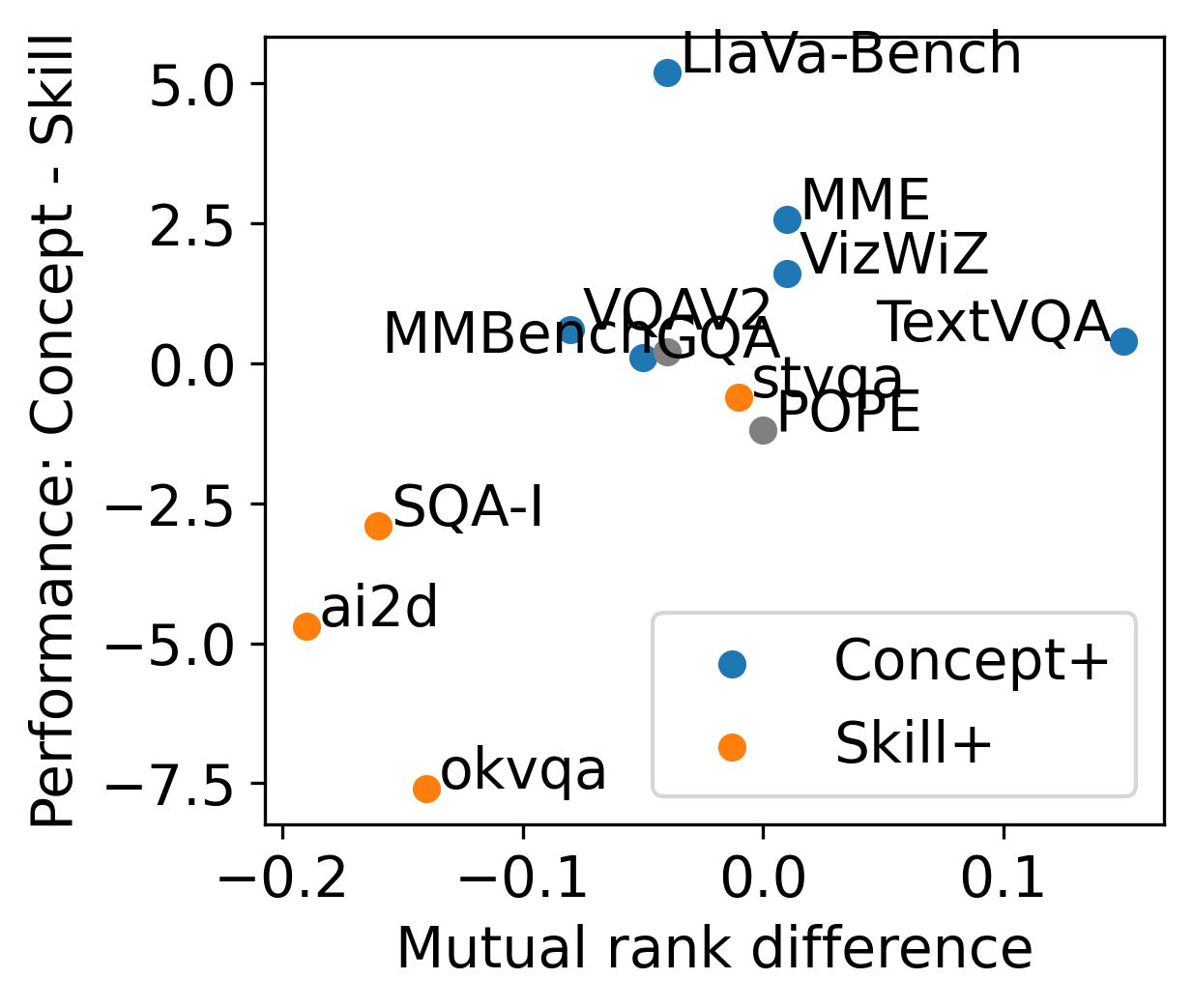}
  \caption{Scatter of mutual rank difference and performance difference when prioritizing concept vs skill neighbors.}
  \label{fig:llava_5pp_mutual_rank_scatter}
\end{figure}

The unsuccessful attempt to achieve the best-of-both-worlds reinforces the importance of being able to predict \emph{which} dimension—concept or skill—is more relevant for a given benchmark. Rather than attempting to combine the two types of selection heuristics, learning to automatically choose between them appears to be a more effective path forward.
We explore whether it is possible to predict which alignment factor dominates a benchmark using a simple mutual ranking analysis. 
The goal is to infer, without running two separate fine-tuning experiments, whether a benchmark is likely to benefit more from concept-targeted or skill-targeted selection.

\paragraph{Method}
For a given benchmark, we first construct two relevance rankings over the entire instruction pool.
The \emph{concept ranking} ranks samples by their visual concept similarity (using concept embedding space similarity) to the benchmark images, and \emph{skill ranking} ranks by their visual skill similarity (using skill embedding space similarity).

To capture the relationship between these two rankings, we examine the top-1 nearest neighbor in each list and measure where that sample appears in the opposite ranking. Concretely, for the top-1 skill-ranked sample, we record its rank in the concept ordering, and for the top-1 concept-ranked sample, we record its rank in the skill ordering. We refer to these two cross-ranks as $R_{c|s}$ and $R_{s|c}$.

\begin{table*}[ht]
    \normalsize
    \centering
    \caption{Skill description comparisons between samples from SQA and their corresponding nearest neighbors in the LLaVA-1.5 training mixture.
    }
    \adjustbox{max width=0.95\linewidth}{
    \begin{threeparttable}
    \begin{tabular}{ll}
        \toprule
        Samples from benchmark &  Nearest neighbors in training set \\
        \midrule
        \midrule
        \makecell[l]{To interpret the graph accurately and identify temperature \\ trends across the months.} &
        \makecell[l]{The ability to interpret and analyze text and numerical data \\ related to temperature ranges.} \\
        \midrule
        \makecell[l]{One must analyze the beak shape of the birds to determine  \\ adaptations for cracking hard seeds.} &
        \makecell[l]{One needs to observe details of the bird's beak shape, \\ feeding behavior, and surrounding environment.} \\
        \midrule
        \makecell[l]{Identifying and comparing the number of pink balls in each  \\ solution.} &
        \makecell[l]{One must identify and differentiate between various types of \\ balls in the image.} \\
        \midrule
        \makecell[l]{The ability to compare the number of seedlings in different \\ pots and analyze growth differences is required.} &
        \makecell[l]{Observation, interpretation of visuals, and understanding of \\ growth requirements for plants.} \\
        \midrule
        \makecell[l]{Identifying organisms based on scientific names and recognizing \\ their taxonomic relationships within a specific context.} &
        \makecell[l]{Identifying the animal's species and recognizing its \\ characteristics for accurate scientific naming.} \\
        \bottomrule
        \end{tabular}
        \end{threeparttable}
        }
    \label{tab:qd_case_study_sqa}
\end{table*}

\paragraph{Interpreting Cross-Ranks}
The cross-ranks reveal how strongly concepts and skills co-occur for a given benchmark:
\begin{itemize}
    \item If $R_{c|s}$ is \emph{low} (i.e., skill neighbors have low concept ranks), this suggests that skill-similar samples are visually diverse, meaning the benchmark emphasizes general reasoning skills rather than specific visual domains.
    \item If $R_{c|s}$ is \emph{high}, it indicates that skill-similar samples also share similar visual concepts, implying that the skills required are tied to particular domains.
    \item Conversely, a high $R_{s|c}$ indicates that concept-similar samples tend to involve similar skills, while a low $R_{s|c}$ indicates that concept neighbors are heterogeneous in skill demands.
\end{itemize}

\paragraph{Analysis}
By comparing these two cross-ranks, we can infer the dominant alignment factor for the benchmark.
\textbf{General-skill benchmarks} tend to have lower $R_{c|s}$ but higher $R_{s|c}$, since general skills apply broadly across many concepts, but concept neighbors mostly correspond to simple questions requiring similar low-level skills.
On the other hand, \textbf{specific-skill benchmarks} tend to have higher $R_{c|s}$ but lower $R_{s|c}$, since samples that are similar in skill often share overlapping visual structures, whereas concept neighbors may not exhibit the specialized skills needed.
This asymmetric pattern allows us to classify a benchmark as being primarily \emph{skill-driven} or \emph{concept-driven} without explicitly running both targeted selection strategies.

\paragraph{Results}
Fig~\ref{fig:llava_5pp_mutual_rank_scatter} shows a scatter plot of the mutual rank difference ($R_{s|c} - R_{c|s}$) on the x-axis and the performance difference of model prioritizing concept vs skill on the y-axis. 
Data points closer to the lower-left correspond to the skill-targeted benchmarks, while the upper-right correspond to the concept-targeted.
The simple cross-ranking heuristic successfully predicted the preferred alignment type, enabling an automated and lightweight benchmark-aware instruction selection policy. 
We find that this predictive approach is more consistent than naively combining the two strategies (as discussed in the previous subsection), since it explicitly identifies which factor—concept or skill—dominates the benchmark’s data distribution.




\subsection{Skill description qualitative study}
To verify that the proposed skill extraction pipeline captures meaningful and non-trivial information, we examined skill descriptions from the SQA-I benchmark and compared them to those of their nearest neighbors in the LLaVA-1.5 training mixture (see Table~\ref{tab:qd_case_study_sqa}). The retrieved neighbors demonstrate that the skill-based representation aligns closely with the reasoning steps needed to solve each question, beyond simple object counting or recognition.

For example, skills such as ``interpreting graph trends across months'' and ``understanding growth requirements of plants from visual cues'' highlight the kinds of subtle, context-dependent details that are not apparent from the image alone. This sanity check suggests that the skill embeddings are effective at grouping instructions by the type of reasoning required, rather than by surface-level visual similarity.

\section{Conclusion}

In this work, we investigated the problem of vision-language instruction selection, with a focus on how selection strategies affect downstream performance.
We introduced targeted instruction selection methods based on visual concepts and skills, and demonstrated that different benchmarks exhibit distinct preferences for one type over the other.
Our experiments revealed that concept- and skill-targeted selection can outperform untargeted baselines on tasks aligned with their respective focuses, particularly under limited instruction budgets. 
Furthermore, a simple mutual-ranking analysis of concept- and skill-based neighbors reveals clear asymmetric patterns that indicate whether a benchmark is concept-driven or skill-driven. 
This insight points toward a lightweight and automatic way to choose the right selection strategy without costly trial-and-error.

\section*{Limitations} 
While our findings highlight the value of targeted instruction selection, several limitations remain. 
First, our taxonomy of visual concepts and skills was derived through a combination of heuristics and semi-automatic labeling, which may not fully capture the nuance or interdependence of multimodal abilities. 
Second, our approach assumes access to benchmark-level information during selection, which may not necessarily hold in deployment scenarios involving novel or private tasks. 
Additionally, our method currently treats each benchmark in isolation and does not account for multi-task generalization or interference effects when tuning a model jointly on instructions selected for multiple tasks. 
Lastly, the effectiveness of our strategies may vary with model size and pretraining quality, and further investigation is needed to evaluate their transferability across architectures.

In future work, we aim to develop automatic methods to infer benchmark alignment preferences without supervision, enabling dynamic, zero-shot selection for unseen tasks. 
We also plan to explore more fine-grained selection criteria that consider interaction effects between concept and skill attributes, and to extend our framework to multitask and continual instruction tuning settings.

\bibliography{custom}

\appendix

\section{Prompt for Extracting Skill Descriptions}
\texttt{Here is a list of questions about an image: \\ \
[QUESTION\_1] \\ \
[QUESTION\_2] \\ \
[QUESTION\_3] \\ \
\ 
Don't answer the above questions directly. What visual skills are required to answer these questions? Answer in one short sentence with less than 20 words without any extra reasoning.}

\section{Evaluation Details}
We follow the evaluation pipeline of COINCIDE~\cite{lee2024concept} for the overlapped benchmarks.
The only minor difference is that we adopted \texttt{gpt-4o-mini} for judging the results for LLaVA-Bench.
We also calculate the accuracy of VizWiZ by excluding the unanswerable subset. 
We found that the performance of the unanswerable subset is better for the worse models, likely because the models output ``unanswerable'' as the response to every question.
For the new benchmarks, we directly utilize the implementation provided from the LMMS-Eval library~\citep{lmms_eval2024}.

\section{Baseline Reproduction Details}
We reproduced the results for both Coincide~\citep{lee2024concept} and PreSel~\citep{safaei2025filter}.
Reproducing Coincide is straightforward since they fully open-sourced the codebase, and their selection method supports selecting an arbitrary number of samples.
PreSel requires first training a reference model on a random 5\% subset of the dataset and then performs selection relying on the signal provided by the reference model.
It is unfair to compare with PreSel at 5\% directly since it would just be 5\% of random samples.
Therefore, the 5\% baseline of PreSel was implemented by selecting an additional 5\% of data with the reference model, effectively utilizing 10\% of the data.
However, PreSel fails to outperform other methods even with this advantage.

\section{Additional Qualitative Study on Skill Descriptions}
\begin{table*}[h!]
    \normalsize
    \centering
    \caption{Skill description comparisons between samples from OK-VQA and their corresponding nearest neighbors in the LLaVA-1.5 training mixture.
    }
    \adjustbox{max width=0.95\linewidth}{
    \begin{threeparttable}
    \begin{tabular}{ll}
        \toprule
        Samples from benchmark &  Nearest neighbors in training set \\
        \midrule
        \midrule
        \makecell[l]{One must recognize the vehicle type and its design characteristics\\ to determine its historical context and invention date.} &
        \makecell[l]{Identifying the vehicle type and recognizing its historical \\ context.} \\
        \midrule
        \makecell[l]{One needs to identify the meal’s ingredients, presentation \\ style, and cultural context to suggest a suitable side dish.} &
        \makecell[l]{One must identify cultural elements in the dish's presentation, \\ ingredients, and style.} \\
        \midrule
        \makecell[l]{Identifying the pastry type, size, and any visible toppings \\ or fillings.} &
        \makecell[l]{One must identify colors, textures, and shapes of the filling \\ in the pastry.} \\
        \midrule
        \makecell[l]{One must identify size, shape, and features typical of buses \\ versus vans to answer the question.} &
        \makecell[l]{You need to identify and differentiate types of buses based \\ on visual characteristics.} \\
        \midrule
        \makecell[l]{Identifying logos, labels, and packaging design details in \\ the image to determine the origin of the beverage.} &
        \makecell[l]{ The ability to identify brand logos and labels on beverage \\ packaging.} \\
        \bottomrule
        \end{tabular}
        \end{threeparttable}
        }
    \label{tab:qd_case_study_okvqa}
\end{table*}

\end{document}